\documentclass{article}

\usepackage[preprint]{neurips_2019}

\usepackage[utf8]{inputenc} 
\usepackage[T1]{fontenc}    
\usepackage{hyperref}       
\usepackage{url}            
\usepackage{booktabs}       
\usepackage{amsfonts}       
\usepackage{nicefrac}       
\usepackage{microtype}      
\usepackage{graphicx}
\usepackage{bm}
\title{Learn to Segment Organs with a Few Bounding Boxes}

\author{%
  Abhijeet Parida\thanks{Authors contributed equally}\\
  Technical University of Munich\\
  \texttt{abhijeet.parida@tum.de} \\
   \And
   Arianne Tran*\\
 Technical University of Munich\\
  \texttt{arianne.t@tum.de} \\
   \And
   Nassir Navab \\
 Technical University of Munich\\
  \texttt{nassir.navab@tum.de} \\
  \And
   Shadi Albarqouni \\
 Technical University of Munich\\
  \texttt{shadi.albarqouni@tum.de} \\  
}

\begin{document}

\maketitle

\begin{abstract}
Semantic segmentation is an import task in the medical field to identify the exact extent and orientation of significant structures like organs and pathology. Deep neural networks can perform this task well by leveraging the information from a large well-labeled data-set. This paper aims to present a method that mitigates the necessity of an extensive well-labeled data-set. This method also addresses semi-supervision by enabling segmentation based on bounding box annotations, avoiding the need for full pixel-level annotations. The network presented consists of a single U-Net based unbranched architecture that generates a few-shot segmentation for an unseen human organ using just 4 example annotations of that specific organ. The network is trained by alternately minimizing a nearest neighbor loss for prototype learning and a weighted cross-entropy loss for segmentation learning to perform a fast 3D segmentation with a median score of 54.64\%.

\end{abstract}

\section{Introduction}
Semantic Segmentation, reported by \citet{Hesamian2019}, is used in the medical field to identify the exact shape and size of structures like organs, and pathology. Deep Learning based image segmentation using the information from large-scale fully-annotated datasets is now a robust tool for medical applications.
The creation of labels for semantic segmentation is, however, a tedious task, requiring annotation of each pixel belonging to a class. According to \citet{3d-few-shot}, this is particularly more cumbersome in the medical domain. The annotation in the medical domain is done by highly specialized experts, who have developed their skills through years of practice. The number of man-hours spent is enormous and most of the time the output is not even optimal due to inter/intra-observer variability. 
In natural images, \citet{COCO} suggests the creation of bounding box labels is 15 times faster and cheaper compared to full pixel label annotations for the same image. This cost gap in the medical domain would, therefore, be even wider given the required expertise.

The few-shot learning (FSL), for semantic segmentation in computer vision, has been thoroughly studied by \citet{shaban,rakelly}, and \citet{BMVC}. All previous approaches, however, suggest a branched structure, with one branch extracting meta-information from the examples, so-called support set, and the other branch uses the meta-information to segment the required image, so-called query image. \citet{shayan} follow the same approach, with the help of Squeeze-and-Excite blocks (\cite{hu2018squeeze}), to segment organs in contrast-enhanced CT scans.

In this paper, we first introduce the prototype learning into a single un-branched architecture for few-shot segmentation reducing the cost of both computation and annotation. Further, we extend our proposed method to the semi-supervised few-shot learning (SS-FSL) setup to leverage weak support annotations, i.e. bounding boxes, which is the first to be seen in few-shot approaches for  medical applications.

\section{Methodology}
Our approach is based on prototype learning, which takes images and corresponding annotations as input and outputs a prototype. The prototype along with the feature representation of the query images are passed then to the model to predict the segmentation maps. Contrary to the branched networks seen in the Few-Shot Learning (FSL) paradigm, both the prototype learner and the segmenter are integrated into a single network. In addition, weakly annotated images, i.e. bounding boxes, are leveraged during the meta-learning phase.

Concretely, for an $N$-shot $K$-way segmentation problem, a model $f(\cdot)$ is trained on episodes $E$ consisting of a support set $\mathcal{S} = \{(x_1, y_1),\cdots,(x_S, y_S)\}$ where each $x_i$ is an input instance image and $y_i$ is the corresponding annotation of $K$ classes, and a query set $\mathcal{Q} = \{x_1, \cdots, x_Q\}$. The model then transfers the knowledge, acquired on the support set $\mathcal{S}$ during the meta-learning, to predict the segmentation maps on the query set $\mathcal{Q}$.
In the semi-supervised FSL setup, we assume access to dataset $\mathcal{D}_{train} =\{\mathcal{D}_L$, $\mathcal{D}_W\}$, where $\mathcal{D}_L$, $\mathcal{D}_W$ denote all fully annotated, and weakly annotated images, respectively, and $\mathcal{D}_{k,L}$ denote all annotated images of a single organ $x \in class (k)$, and $\mathcal{D}_{k,W}$ be all weakly annotated images $x \in class (k)$.
In each training episode $E_k$, both support $\mathcal{S}_k$ and query sets $\mathcal{Q}_k$ are sampled, for a particular class $k$, i.e. organ, from both $\mathcal{D}_{k,L}$, and $\mathcal{D}_{k,W}$ with different proportion, where the support $|A|$ of annotated samples is way larger than the support $|B|$ of weakly annotated ones, i.e. $B >> A$.

\textbf{Prototype Learner:} Both  an image $x_{k}^{Q} \in Q_k$ and corresponding annotation $y_{k}^{S}$ are passed as input to form a prototype $\boldsymbol{\hat{p}}_{k}=f_\theta(x_{k}^{Q} ,y_{k}^{S})$, which is later passed as a meta information to the segmenter. The model parameters $\theta$ are optimized by minimizing the negative log likelihood, 
\begin{equation}
\mathcal{L}_{NN}(\theta)=-\log\left(\frac{\exp(d(\boldsymbol{\hat{p}}_{k},\boldsymbol{p}_{k}))}{\sum_{k^{\prime}=1}^{K}\exp(d(\boldsymbol{\hat{p}}_{k},\boldsymbol{p}_{k\prime}))}\right), 
\label{eqn:proto2}
\end{equation}
where $d(\cdot, \cdot)$ is a similarity function, i.e. cosine similarity, and $\boldsymbol{\hat{p}}_{k}$ is the prototype computed on the support set, $\boldsymbol{p}_{k} = \sum_{\mathcal{S}_k} f_\theta(x_{k}^{Q} ,y_{k}^{S})$. The Nearest Neighbour Loss($\mathcal{L}_{NN}$) assists in the formation of individual clusters for all $class (k)$.

\textbf{Segmenter:} The learned prototype along with the query images $x^Q_{k}$ are passed to the model to predict the segmentation maps, $\hat{y}_k=f_{\phi, \theta}(x^Q_{k},y^S_k)$. The model parameters $\phi$, and $\theta$ are optimized using the weighted cross-entropy loss, 
\begin{equation}
\mathcal{L}_{WCE}(\phi, \theta)=-\beta \dot y_k \log (\hat{y_k})+(1-y_k) \log (1-\hat{y_k})), 
\label{eqn:wce}
\end{equation}
where $\beta$ is additional weight to balance the the occurrence of class $k$. 

The overall loss function is minimized alternatively to enable learning from a few shot segmentation, 
\begin{equation}
\mathcal{L}(\phi, \theta)= \mathcal{L}_{NN}(\theta) + \mathcal{L}_{WCE}(\phi, \theta). 
\label{eqn:total}
\end{equation}

\section{Experiments and Results}
We evaluate our proposed model on the publicly available Visceral dataset created by~\citet{viseral}. We follow the same evaluation strategy that appeared in \citet{shayan}by focusing on 4 organs. 
Each organ class has a data set of 20 different patients, of which we use 19 for the query set and the remaining one to provide the supporting meta information. We obtain the results by following a fold wise training strategy. A fold is defined by treating one of the organs as an previously unseen testing class $D_{test}$ while utilizing the rest of the organs as $D_{train}$ for training. 
We train our network with the aforementioned setting both in a fully supervised fashion and a semi-supervised fashion by using weakly labeled data, i.e. bounding box labels.

\begin{figure}[h!]
  \centering
  \includegraphics[width=0.95\textwidth]{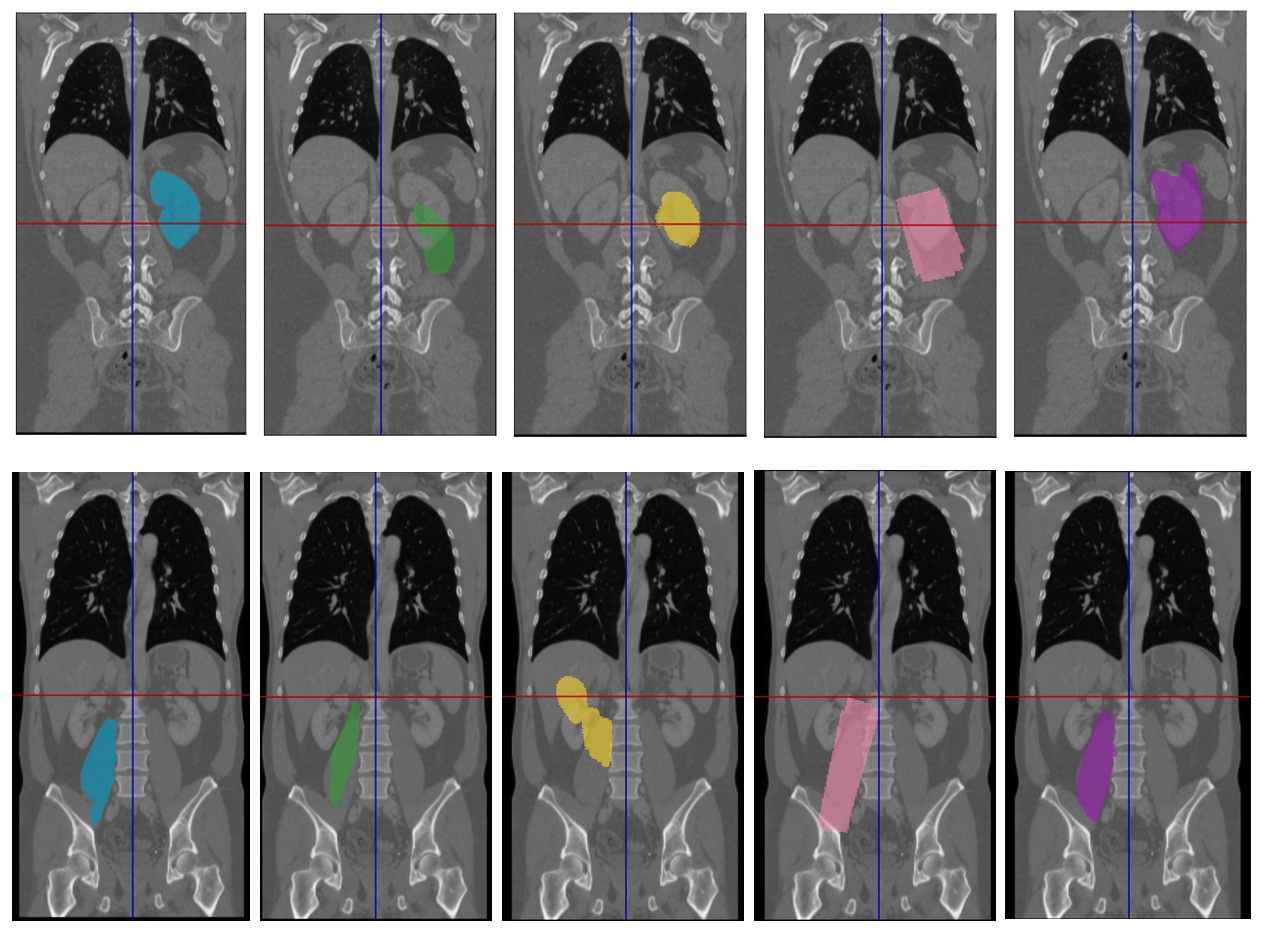}
  \caption{The upper row of the figure displays segmentation results for the left kidney, the lower row results for the right psoas major. The blue label is the ground truth, green the support input of the fully supervised, yellow the segmentation result of the fully supervised, pink the support input of the semi-supervised and purple is the segmentation result of the semi-supervised case.}
  \label{fig:organs1}
\end{figure}

In Fig.~\ref{fig:organs1}, we can see that even though the support label for the fully supervised case can be considered closer to the ground truth, the result for the vague box-shaped support label yields an overall better performance in terms of results, showing an increase of the 3D dice score of around 10\% on average for each organ. Another aspect is the number of used ground truth, compared to conventional full-shot techniques, we yield an annotation cost decrease of almost 500x for the fully supervised few-shot case and even up to 7500x decrease for the bonding box semi-supervised case.
We were able to prove that semi-supervised few-shot learning (SS-FSL) contributes a valuable improvement to few-shot techniques, as shown in Table~\ref{tab:result-table}.
Even though we do not reach the performance of state-of-the-art methods as full shot techniques, our approach has a great potential due to these attributes. 

\begin{table}[t]
  \caption{The volumetric dice coefficient of several organs and the speedup performance of our method compared to the Lower Bound Model (LBM), i.e. fully-supervised FSL, and Upper Bound Model (UBM), i.e. fully supervised models.
}
  \label{tab:result-table}
  \centering
  \begin{tabular}{llcccccc}
    \toprule
             & & 
    \multicolumn{4}{c}{Organs}                   &\\
     \cmidrule(r){3-6} 
     &   & Liver & Spleen& Kidney& Psoas Major    & Average & Speedup\\
    \midrule
LBM:FSL &    Ours& 52.78&53.09 &\textbf{50.52} & 35.82   &48.06 & \textbf{2.183}\\
    & \citet{shayan}&\textbf{68.04} &55.06 &37.62 &\textbf{49.92}    &52.66 & 1 \\
\midrule
    SS-FSL & Ours& 63.50&\textbf{64.28} &45.75 & 45.01   &\textbf{54.64}& \textbf{2.183} \\
\midrule
    UBM& INSA-Creatis &95.11 & 91.10&95.00 &81.20    &90.61 & -- \\
    \bottomrule
  \end{tabular}
\end{table}

\section{Conclusion}
We present a method for segmenting 3D medical scans in a low data setting. This approach can be adapted to enhance any segmentation network leaving room for improvement and further research.
Our proposed SS-FSL yielded an increase of 10\% performance organ-wise for most of the organs compared to full supervision.  Few-Shot Learning is an interesting technique with lots of potential in the medical domain where there is a lack of good reliable large data-sets.

\bibliographystyle{plainnat} 
\bibliography{literature}

\newpage
\appendix
\section{Appendices}
\subsection{Implementation}

We base our model on U-Net~\citet{UNET2}, the most widely used segmentation architecture in medical imaging.

The architecture is provided with the required meta-information at every
encoding stage, except at the first one. The information is supplied from the incoming data by simply doing an element-wise multiplication of the extracted features with the meta-information. The meta-information comes from the support annotation of the same organ from another patient. The support information is obtained from 1 fully labelled and 3 weakly labelled annotations.

The training is done in two phase joint training described by \citet{BMVC}. In the first phase, the query image and the support information leads to obtaining a 1024 dimensional bottleneck features. These features are passed through a 1x1 convolutional layer which brings down its
dimensions from 1024 to 64. The 64-dimensional feature vector is then passed through a Global Average Pooling (GAP) suggested by \citet{GAP}. This averages out the features in a channel to obtain the prototypes. Further \citet{BMVC} point out that the GAP in contrast to Max or Min pooling helps prevent over-fitting to dominant or submissive features of the training organ. This obtained prototype is used
to minimize the nearest neighbor loss with the learned standard prototypes of its class.

\begin{figure}[h!]
  \centering
  \includegraphics[width=0.9\textwidth]{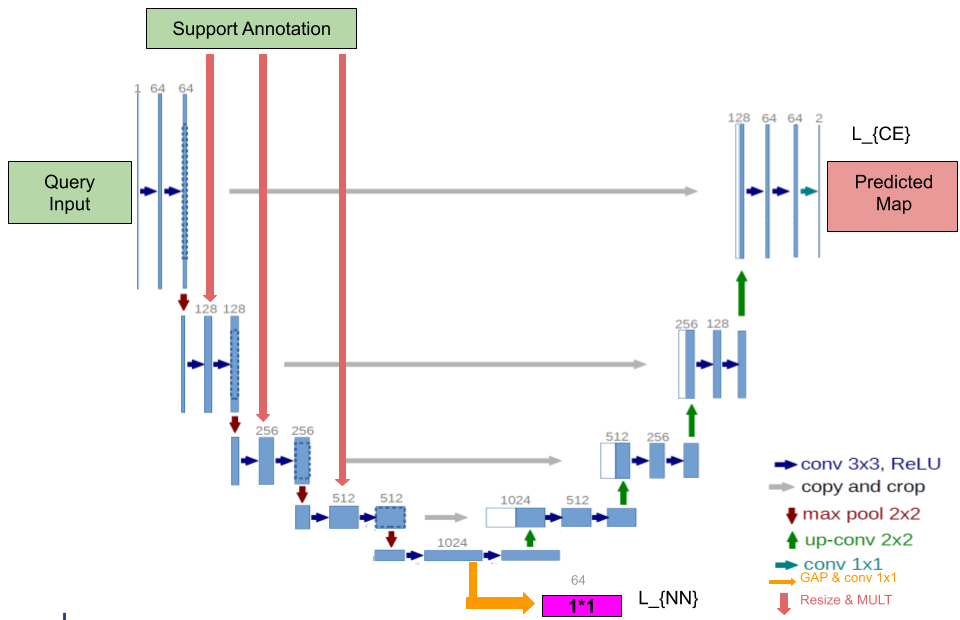}
  \caption{A schematic of the UNet architecture adapted from \citet{UNET} used for few shot training with the semi supervised support annotation. In the figure , the query is the input slice that needs to be segmented for a particular organ. The support annotation consist of an examples of the particular organ(previously unseen organ during testing). The pink box is the prototype that has been been learned by minimizing a nearest neighbour loss. The predicted map is the segmentation produced for the organ in the query input for which the support has examples. The segmentation map is learned by minimizing the the weighted cross entropy loss}
  \label{fig:unet}
\end{figure}

In the second phase, a contextual information  is learned to segment the query. The same query and support are again passed through to optimize for the cross-entropy for the segmentation accuracy. The entire schematic for the two-phase training is given in Fig. \ref{fig:unet}.

UNet's symmetric expanding path enables precise localization. The symmetric expansion path with the skip connection provides the decoder with boundary information from the input image to create the final segmentation mask.
The idea of masking support annotations at every encoding layer may prevent the carrying of boundary related information to the decoding layer. The information may be preserved by simply not masking the support at the very first encoding block. This also serves to provide the related boundary to the very end of decoding making the data relevant to the predicted map.

\end{document}